\documentclass[10pt, a4paper]{article}
\usepackage{lrec}
\usepackage{graphicx}
\usepackage{tabularx}
\usepackage{soul}
\usepackage{epstopdf}
\usepackage[latin1]{inputenc}
\PassOptionsToPackage{hyphens}{url}\usepackage{hyperref}
\usepackage{xstring}
\usepackage{todonotes}

\title{Multilingual Stance Detection: The Catalonia Independence Corpus}

\name{Elena Zotova$^1$, Rodrigo Agerri$^1$, Manuel Nu\~nez$^2$, German Rigau$^1$}

\address{$^1$ IXA Group, HiTZ Centre, University of the Basque Country UPV/EHU, Donostia-San Sebastian, Spain\\
$^2$ Intercom Strategys, Madrid, Spain\\
zotova.el@gmail.com, rodrigo.agerri@ehu.eus, manuel.nunez@lunigtuk.com,
german.rigau@ehu.eus}

\abstract{Stance detection aims to determine the attitude of a given text with respect to a specific topic or claim. While stance detection has been fairly well researched in the last years, most the work has been focused on English. This is mainly due to the relative lack of annotated data in other languages. The TW-10 Referendum Dataset released at IberEval 2018 is a previous effort to provide multilingual stance-annotated data in Catalan and Spanish. Unfortunately, the TW-10 Catalan subset is extremely imbalanced. This paper addresses these issues by presenting a new multilingual dataset for stance detection in Twitter for the Catalan and Spanish languages, with the aim of facilitating research on stance detection in multilingual and cross-lingual settings. The dataset is annotated with stance towards one topic, namely, the independence of Catalonia. We also provide a semi-automatic method to annotate the dataset based on a categorization of Twitter users. We experiment on the new corpus with a number of supervised approaches, including linear classifiers and deep learning methods. Comparison of our new corpus with the with the TW-1O dataset shows both the benefits and potential of a well balanced corpus for multilingual and cross-lingual research on stance detection. Finally, we establish new state-of-the-art results on the TW-10 dataset, both for Catalan and Spanish.\\ \newline  }

\begin{document}

\maketitleabstract

\section{Introduction}

The rise of social media has given rise to the ``fake news'' phenomenon. According to the Fake News Challenge, ``Fake news, defined by the New York Times as ``a made-up story with an intention to deceive''\footnote{https://www.nytimes.com/2016/12/06/us/fake-news-partisan-republican-democrat.html}, often for a secondary gain, is arguably one of the most serious challenges facing the news industry today.''\footnote{http://www.fakenewschallenge.org/}

Determining the veracity of a given document or story, namely, whether it is fake or legitimate, is a very complex task, even for expert fact-checkers. Thus, previous work breaks down the fake news detection task in different stages, the first of which is establishing what other news sources are saying about the given document or story (whether they agree, disagree, etc. with the news story), namely, determining their stance with respect to that document or news story. Following this, the first stage of the Fake News Challenge was \emph{Stance Detection}. This decision was supported by two main ideas: (i) a stance detection system should allow a human fact checker to enter a document (headline, message, claim, etc.) and retrieve the top documents from other news sources that agree, disagree or discuss the given document and, (ii) based on the previous step, it would be possible to build a ``truth-labeling'' system based on the weighted credibility of the various news organizations from which the stance has been retrieved.

Automatic stance detection has been defined as the task of classifying the attitude expressed in a text towards a given target or claim. Most of the work on stance detection has been undertaken in English using the data provided by the Detecting Stance in Tweets shared task organized at SemEval 2016 \cite{mohammad-etal-2016-semeval}, RumourEval 2017 \cite{derczynski-etal-2017-semeval} and the Fake News Challenge. The SemEval 2016 task was formulated as follows: given a tweet text and a target entity or topic, automatic natural language systems must determine whether the tweet expresses a stance in \textbf{favor} of the given target, \textbf{against} the given target, or whether \textbf{none} of those inferences are likely. For example, consider the following target$-$tweet pairs:

\begin{quote}
\textbf{Tweet:} \textit{I still remember the days when I prayed God for strength.. then suddenly God gave me difficulties to make me strong. Thank you God! \#SemST}

\textbf{Target:} Atheism

\textbf{Stance:} AGAINST

\vspace{0.5cm}

\textbf{Tweet:} \textit{@PH4NT4M @MarcusChoOo @CheyenneWYN women. The term is women. Misogynist! \#SemST}

\textbf{Target:} Feminist Movement

\textbf{Stance:} FAVOR
\end{quote}

These examples illustrate the nature of the task, in which tweets are very short, full of specific vocabulary, non-standard spelling grammar, emojis, hashtags, and high on irony and sarcasm. The task aimed to detect stance from single tweets, without taking into account the conversational structure of tweet threads or any information about authors.

Following the model of the SemEval 2016 initiative, two shared tasks were organized as part of IberEval workshop \cite{taule17,taule18}. They provided tweets annotated for Stance in Catalan and Spanish. The target of the 2017 edition was the ``Catalan Independence'' whereas the 2018 edition (TW-10 dataset) focused on the ``Catalan referendum on the 1st of October''. In both editions the classes distribution was hugely skewed, which makes it difficult to explore and compare stance detection methods in multilingual and cross-lingual settings.

In this context, we propose the new Catalan Independence Corpus (CIC) for stance detection in Catalan and Spanish. By doing so, we aim to promote research in other languages different to English. Furthermore, the corpus presents a balanced distribution between classes so that researchers can explore multilingual and cross-lingual methods.

The contributions of this paper are the following: (i) we present a new dataset in Catalan and Spanish to work on multilingual and cross-lingual stance detection; (ii) we propose a semi-automatic method to collect and annotate a corpus of tweets based on a categorization of Twitter users. This method partially alleviates the huge effort of manually annotating the corpus tweet by tweet; (iii) we report new state-of-the-art results on the TW-10 dataset of IberEval 2018 \cite{taule18}; (iv) comparison between results using our new corpus and the TW-10 dataset shows the benefits of providing a balanced multilingual corpus, and (v) both the datasets and code are made public to facilitate future research and reproducibility of results\footnote{https://github.com/ixa-ehu/catalonia-independence-corpus}.

\section{Related Work}\label{sec_related_work}

The state of the art is divided into two main approaches. First, those that rely on \emph{traditional} machine learning models combined with hand-engineered features \cite{Mohammad:2017:SST:3106680.3003433} or vector-based word representations (word embeddings) \cite{bohler-etal-2016-idi}. In particular, \cite{Mohammad:2017:SST:3106680.3003433} obtained the best results for the supervised setting of the SemEval 2016 task using a SVM classifier to learn word n-grams (1-, 2-, and 3-gram) and character n-grams (2-, 3-, 4-, and 5-gram) features, outperforming deep learning approaches \cite{zarrella-marsh-2016-mitre,wei-etal-2016-pkudblab}.

Among the deep learning systems published, the pkudblab system \cite{wei-etal-2016-pkudblab} proposed a Convolutional Neural Network (CNN) architecture combined with a voting scheme to guide the predictions instead of generating them based on the accuracy obtained in the validation set. The MITRE team \cite{zarrella-marsh-2016-mitre} employed two recurrent RNN classifiers: the first was trained to predict task-relevant hashtags on a large unlabeled Twitter corpus which was then used to initialize a second RNN to be trained on the SemEval 2016 training set. \cite{du2017stance} proposed a neural network-based model to incorporate target-specific information by means of an attention mechanism. Finally, \cite{sun-etal-2018-stance} proposed a hierarchical attention network to weigh the importance of various linguistic information, and learn the mutual attention between the document and the linguistic information.

It should be said that neural network approaches have been more successful so far for the SemeEval 2016 Task B (weakly-supervised setting). Apart from the previously mentioned systems \cite{wei-etal-2016-pkudblab}, \cite{augenstein-etal-2016-stance} proposed a bidirectional Long-Short Term Memory (LSTM) encoding model. First, the target is encoded by a LSTM network and then a second LSTM is used to encode the tweet using the encoding of the target as its initial state.

Another interesting work is that of \cite{DBLP:journals/corr/RajadesinganL14} who tried to determine stance at user level. Their assumption was that if many users retweeted a particular pair of tweets in a short time, then it is likely that this pair of tweets had something in common and share the same opinion on the topic.

As far as we know, most approaches to stance detection are developed for English, with the few exceptions that use the Catalan and Spanish data from IberEval 2017 and 2018 \cite{taule17,taule18} or the work of \cite{mohtarami-etal-2019-contrastive} using the Arabic corpus provided by \cite{baly-etal-2018-integrating}.

With respect to the ``MultiModal Stance Detection in tweets on Catalan \#1Oct Referendum'' task at IberEval 2018 (MultiStanceCat), the best results for Spanish were obtained by the uc3m team \cite{Segura-Bedmar18}. They presented a system based on bag-of-words with TF-IDF vectorization. They evaluated several of the most commonly used classifiers, obtaining a final 28.02 F1 macro score in the Spanish test data. The best result in Catalan subset was obtained by the {CriCa} team \cite{Cuquerella2018CriCaTM}. Their approach consisted of combining the Spanish and Catalan subsets to create a larger and more balanced corpus. They experimented with stemming of various lengths (three, four and five characters) and removing character suffixes from the word. Since Spanish and Catalan share many words, stemming helped to generalize. Additionally, it is quite common to encounter tweets containing words and expressions in both languages. Their final F1 macro was 30.68.

\section{Experimental Setup}\label{sec:experiment}

The development of the Catalonia Independence Corpus was motivated by the experiments performed on the IberEval TW-10 data. The result of those experiments showed that, due to the highly imbalanced nature of the TW-10 corpus, any comparison of systems across languages were not particularly meaningful. In this section we will summarize the setup for the experiments performed on both datasets, TW-10 and our new Catalonia Independence Corpus.

Apart from the data pre-processing described in Section \ref{sec:data-pre-processing}, we experimented with four different system architectures: (i) TF-IDF vectorization with a SVM classifier; (ii) a SVM trained with fastText word embeddings \cite{Grave18} for the representation of tweets; (iii) the fastText text classification system \cite{joulin-etal-2017-bag} with fastText word embeddings and, finally (iv) the Flair system \cite{akbik-etal-2018-contextual}, which implements a Recurrent Neural Network (RNN) for text classification that can be combined with static and context-based string embeddings. In the following, we describe the pre-processing and each of the architectures tested in both the TW-10 and the Catalonia Independence Corpus (CIC).

\subsection{Data Pre-processing}\label{sec:data-pre-processing}

Since each tweet in the TW-1O dataset is given in context, with the previous and the next tweet, we use them to obtain longer and richer texts for classification.

\textbf{Normalization}: We believe that normalization helps to reduce the number of features for TF-IDF feature representation and to maximize the number of words that correspond with the vocabulary of pre-trained word vector models. First, we remove all punctuation and any expression starting with "@", "RT", URLs and numbers. The next step is lowercasing and normalization of spelling: we remove repeated characters with one and replacing common shortened words to their normal form. For example, \textit{holaaaaaaa} is converted to \textit{hola}. However, we leave untouched consonants composed of two characters (\emph{tt, ll, rr}). Finally, diacritics are systematically removed.

\textbf{Lemmatization}: Next, we apply a simplified version of lemmatization consisting of replacing the word form with its lemma via dictionary look-up \footnote{https://github.com/michmech/lemmatization-lists}. If a word is not found we leave it in its original form. Note that this method is not capable of resolving ambiguities. For example, the Spanish preposition \textit{para} (``for'') and the verb \textit{para} (``stop'') will be mapped to the same lemma, namely, \textit{parar} (the infinitive ``to stop'' in Spanish). Furthermore, named entities are sometimes wrongly lemmatized. To reduce the error rate, we manually edited the list of lemmas, and deleted the less frequent ambiguous words. In any case, our experiments showed that this type of lemmatization reduces dramatically the number of features helping to improve results for every experimental setting. In addition, it allows to deal with unseen words. For example, if the Spanish word \textit{andando} (walking) does not appear in the training corpus but another form of its lemma does, then both words will be recognized as having the same lemma, namely, the Spanish verb \textit{andar} (to walk).

\textbf{Tokenization}: We perform whitespace tokenization, also removing stopwords (auxiliary verbs, prepositions, articles, pronouns and the most frequent words) and words shorter than three characters.

\subsection{SVM+TF-IDF}\label{sec:tf-idf+svm}

\textbf{TF-IDF} (Term Frequency times Inverse Document Frequency) \cite{Jones72astatistical} is a weighting scheme broadly used in many tasks. Its goal is to reduce the impact of words that occur too frequently in a given corpus. TF-IDF is the product of two metrics, the term frequency and the inverse document frequency.
We calculate the TF-IDF scores for all pre-processed unigrams in the training corpus. The number of features equals the size of the vocabulary of the dataset and represents the dimensionality of the document vector.

\textbf{Information Gain} is used for feature selection \cite{Cover:2006:EIT:1146355}. Information Gain provides a method to calculate the mutual information between the features and the classification labels. According to \cite{Aggarwal12}, mutual information is defined on the basis of the level of co-occurrence between the label and word. In other words, it represents the predictive power of each feature, and measures the number of bits of information obtained for prediction of a class in terms of the presence or absence of a feature in a document. The Information Gain scores show how common a specific feature is in a target class. For example, those words that occur mainly in tweets labelled as FAVOR will be highly ranked. All the weights are normalized and the features ranked from one to zero. We then select those features that are larger than zero.

\textbf{Grid Search} is performed for hyper-parameter optimisation. The grid-search results are measured by 5-fold cross-validation on the training set. To reduce the cost of the grid-search process, we select two of the SVM (RBF kernel) parameters, namely, C and gamma.

\subsection{SVM+fastText Embeddings}\label{sec:svm+f-embedd}

Word embeddings encode words as continuous real-valued representations in a low dimensional space. Word embeddings are trained over large corpora and are able to capture semantic and syntactic similarities based on co-ocurrences. Word embeddings allow to build rich representations of text and have enabled improvements across most NLP tasks.

To the best of our knowledge, the only publicly available pre-trained models for both Catalan and Spanish are those distributed by fastText \cite{Grave18}. Initial experimentation showed that the Common Crawl\footnote{http://commoncrawl.org/} models performed better for our particular task. The Common Crawl models are trained using a Continuous Bag-of-Words (CBOW) architecture with position-weights and 300 dimensions on a vocabulary of 2M words. In order to produce vectors for out-of-vocabulary words, fastText word embeddings are trained with character n-grams of length 5, and a window of size 5 and 10 negatives \cite{Grave18}. We represent the tweet as the average of its word vectors \cite{DBLP:journals/corr/KenterBR16}, which is calculated as follows:

\[V(t)=\frac{1}{n}\sum_{i=1}^{n}W_{i}\]

where \textit{V(t)} is the vector representing a tweet, \textit{n} is the number of words and \textit{W} the vector for each word. In order to facilitate the look-up into the pre-trained word embedding model, the pre-processing described in the previous section is modified, leaving untouched the diacritics and the stopwords.

\subsection{FastText System}\label{sec:fasttext-system}

Apart from the pre-trained word embedding models, fastText also refers to a text classification system \cite{joulin-etal-2017-bag}. The fastText system consists of a linear model with rank constraint. A first weight matrix A is build via a look-up table over the words. Then the word representations are averaged to construct the tweet representation, which is then fed into a linear classifier. This is similar to the previous approach, but in the fastText system the textual representation of the tweet is a hidden variable which can be reused. The CBOW model proposed by \cite{mikolov2013b} is similar to this architecture, with the difference that the middle word is replaced by the stance label. Finally, fastText uses a softmax function to calculate the probability distribution over the predefined classes.

We use the fastText system in its default parameters, with the following exceptions: (i) instead of training the word embeddings online, we provide as input the pre-trained fastText word embedding models for Catalan and Spanish described in the previous section and, (ii) we use bag of bi-grams and trigrams as additional features with the aim of capturing word order information.

\subsection{Neural Architecture}\label{sec:neural-architecture}

Flair refers to both a deep learning system and to a specific type of character-based contextual word embeddings. While fastText generates static word embeddings, generating a unique vector-based representation for a given word independently of the context, contextual word embeddings aimed to generate different word representations depending on the context in which the word occurs. Examples of such contextual representations are ELMo \cite{Peters:2018} and Flair \cite{akbik-etal-2018-contextual}, which are built upon LSTM-based architectures and trained as language models.

The Flair toolkit \cite{akbik-etal-2019-flair} allows to train sequence labelling and text classification models based on neural networks. Flair provides a common interface to use and combine different word embeddings, including both Flair and fastText embeddings. For text classification the computed word embeddings are fed into a BiLSTM to produce a document level embedding which is then used in a linear layer to make the class prediction. For best results, we follow their advice of combining in a stack the contextual Flair embeddings for Spanish with the fastText embeddings \cite{akbik-etal-2018-contextual}. Every result reported with Flair is the average five training runs initialized at random.

\subsection{Evaluation}\label{sec:evaluation}

The models are tuned via cross-validation for the TW-10 dataset. The Catalonia Independence Corpus provides a development set which is used for tuning the models during training. The metric used by the organizers of SemEval 2016 \cite{mohammad-etal-2016-semeval} and IberEval 2018 \cite{taule18} reported the F1 macro-average score of two classes: FAVOR and AGAINST, although the NONE class is also represented in the test data. We use the provided evaluation script \footnote{http://alt.qcri.org/semeval2016/task6/} that calculates the final F1 macro score:

\[F1_{macro} = \frac{F1_{favor} + F1_{against}}{2}\]

\section{TW-1O Referendum Dataset}\label{sec:tw1o_dataset}

The TW-10 for IberEval 2018 dataset was collected using the hashtags \#1oct, \#1O, \#1oct2017 and \#1octl6 to obtain the tweets from Twitter \cite{taule18}. These hashtags were widely used in the debate on the right to hold a referendum on Catalan independence on the 1st of October 2017. A total of 87,449 tweets in Catalan and 132,699 tweets in Spanish were collected between the 20th and 30th of September. The final dataset consists of 11,398 tweets: 5,853 written in Catalan (the TW-1O-CA corpus) and 5,545 in Spanish (the TW-1O-ES corpus). The dataset was annotated manually by three experts. Also, each tweet is given together with its previous and next tweets as context. Table \ref{tab:length_tw_dataset} shows the average length of tweets after concatenating the tweet with its context.

\begin{table}[!ht]
\begin{tabular}{llc} \hline
  TW-10 corpus & Catalan & Spanish \\ \hline
  Average tweet length (tokens) & 37.69 & 38.86\\ \hline
\end{tabular}
\caption{Average length of tweets plus their context in the TW-1O corpus.}\label{tab:length_tw_dataset}
\end{table}






Table \ref{tab:twdatasetdistr} illustrates the imbalanced nature of the Catalan subset, which makes it difficult to built and compare models for Catalan and across languages. Thus, while for Spanish the distribution of classes is quite similar, in Catalan the FAVOR class occurs 35 times more than AGAINST, and 8 times more than NONE.

\begin{table}[!ht]
\centering
\begin{tabular}{lcc} \hline
      Label & Catalan & Spanish\\ \hline
    Against & 120 & 1785 \\
      Favor & 4085 & 1680 \\
     None & 479 & 972 \\ \hline
      Total & 4684 & 4437 \\ \hline
\end{tabular}
\caption{Distribution of classes in the TW-1O trainset.}\label{tab:twdatasetdistr}
\end{table}

Tables \ref{tab:result_tw1o_ca} and \ref{tab:result_tw1o_es} reports our results for Catalan and Spanish respectively. It is clear that the Catalan subset makes it very difficult to perform any meaningful experiments given its class distribution. While the best approach for Catalan is SVM+TDF-IDF, it is clear that the results are heavily influenced by the under-represented AGAINST class.

\begin{table}[!ht]\small
\centering
\begin{tabular}{lccc} \hline
\textbf{System} & F1$_{against}$ & F1$_{favor}$ & F1$_{macro}$\\ \hline
SVM+TF-IDF & 22.86 & 94.68 & \textbf{58.77} \\
SVM+FTEmb & 0.00 & 93.88 & 46.94 \\
fastText+FTEmb & 12.90 & 94.60 & 53.78 \\
Flair+FTEmb & 14.79 & 94.40 & 54.59 \\ \hline
\textbf{Baseline} \\
\scriptsize{\cite{Cuquerella2018CriCaTM}} & - & - & 30.68 \\  \hline
\end{tabular}
\caption{Results on the TW-1O Catalan testset.}\label{tab:result_tw1o_ca}
\end{table}

\begin{table}[!ht]\small
\centering
\begin{tabular}{lccc} \hline
\textbf{System} & F1$_{against}$ & F1$_{favor}$ & F1$_{macro}$ \\ \hline
SVM+TF-IDF & 68.50 & 64.53 & 66.52 \\
SVM+FTEmb & 63.65 & 58.85 & 61.25 \\
fastText+FTEmb & 69.58 & 65.37 & \textbf{67.48} \\
Flair+FTEmb & 60.23 & 52.44 & 56.34 \\ \hline
\textbf{Baseline} \\
\cite{Segura-Bedmar18} & - & - & 28.02 \\ \hline
\end{tabular}
\caption{Results on the TW-10 Spanish testset.}\label{tab:result_tw1o_es}
\end{table}

The results for Spanish are a little bit more interesting. First, we can see that the fastText linear classifier combined with fastText embeddings (fastText+FTEmb) obtains much better results than SVM+FTEmb. As the document representation is the same, that means that the fastText classifier \cite{joulin-etal-2017-bag} improves over the performance of SVM. Finally, our results provide a significant improvement over previous state-of-the-art in this dataset for both languages.

Nonetheless, motivated by the results obtained for Catalan, we decided to propose a new multilingual corpus for stance detection with a better distribution of classes.

\section{Catalonia Independence Corpus 2019}\label{sec:independence}

During the process of developing the Catalonia Independence Corpus (CIC) we tried to address the main shortcomings of the TW-10 dataset, as it has been described in Section \ref{sec:tw1o_dataset}.

We had at our disposal a collection of tweets from 12 days during February and March of 2019 posted in Barcelona and during September 2018 posted in the town of Terrassa, Catalonia, prepared for commercial research in stance detection and political ideology (left-right) prediction. The collection process was performed by crawling with full access to the Twitter API, obtaining messages of up to the official limit of 240 characters. We decided to use it to create a new dataset for academic research. In order to do so, first we separated them by language\footnote{https://code.google.com/archive/p/language-detection/} and obtained 680000 tweets in Catalan and 2 million tweets in Spanish. We then processed each set separately. We discarded tweets with identical messages and tweets containing less than three words.

\textbf{Annotation} was performed using the same three labels and guidelines as the previously described datasets (SemEval 2016 and TW-10). Thus, FAVOR will state a positive stance towards the independence of Catalonia, AGAINST the opposite, and NONE will express neither a negative nor a positive stance, or simply that it is not possible to reach a conclusion.

\subsection{User Categorization}\label{sec:categ-at-user}

Unlike previous approaches, we do not annotate manually each tweet. Instead, the annotation process is based on classifying users. We first compiled a list of Twitter accounts from media, political parties and political activists that clearly and explicitly express their stance with respect to the independence of Catalonia. Secondly, we extracted the most retweeted tweets and categorized their authors manually by checking their Twitter accounts. The assumption was that for a person it is easier to annotate a whole Twitter account rather than the text of a single tweet without context. The decision about their stance was also made taking into account other aspects from the users' accounts, such as the use of special emojis and symbols that may state clearly the stance towards the target (e.g., displaying a yellow ribbon or a Spanish or Republican Catalan flag, etc.), or by the Bio section. We follow this process to assign a FAVOR, AGAINST or NEUTRAL stance to each user.

Furthermore, we extracted the relations between users based on their retweets \cite{SNA2002}. Assuming that all those who make a retweet share the author's opinion, we categorized these users with the same label as the author of the retweeted message. While this method may introduce some noise, it allowed us to quickly obtain a large amount of annotated data quite cheaply.

In total, 25,510 users were categorized. We do not distinguish between Catalan and Spanish tweets because most of the active users in Catalonia are bilingual and can write in both languages. Table \ref{table:users} reports the distribution of the categorized users. The final set contains 131022 unique tweets in Catalan and 202645 unique tweets in Spanish.

\begin{table}[!ht]
\centering
\begin{tabular}{lc} \hline
      Label & Count \\ \hline
      Favor & 22247 \\
      Against & 3091 \\
     Neutral & 176 \\ \hline
\end{tabular}
\caption{Distribution of the categorized users.}\label{table:users}
\end{table}

\subsubsection{Topic Detection}\label{sec:topic-detection}

We annotated the corpus assigning the stance classes to usernames. However, this does mean that we can use every tweet from the users, given that many messages may not be related to the independence of Catalonia. In order to address this issue we performed the following steps:

\textbf{Hashtags and keywords}: We extracted all the hashtags from the corpus and selected manually those that were related to the independence of Catalonia, such as \textit{\#Catalu\~naesEspa\~na, \#CatalanRepublic, \#Tabarnia, \#GolpeDeEstado, \#independ\'encia, \#judicifarsa, \#CatalanReferendum} etc., totalling 450 hashtags. We also added keywords in both languages, 25 in total. We marked each tweet as being on topic if it contained one of the relevant hashtags or keywords. Table \ref{tab:relevanttweets} displays the distribution of tweets after applying the hashtags and keywords filter.

\begin{table}[!ht]
\centering
\begin{tabular}{lcc} \hline
Label & Catalan & Spanish\\ \hline
 Against & 1476 & 8267 \\
Favor & 23030 & 11843 \\
Neutral & 986 & 497 \\ \hline
\end{tabular}
\caption{Distribution of tweets obtained by hashtags and keywords related to ``independence''.}\label{tab:relevanttweets}
\end{table}

\textbf{Topic modelling}: We can see in Table \ref{tab:relevanttweets} that the vast majority of the tweets are labelled as FAVOR. In order to obtain a balanced dataset, we needed to add more tweets to the under-represented classes. We use the MALLET \cite{McCallumMALLET} implementation of Latent Dirichlet allocation (LDA) \cite{Blei:2003:LDA:944919.944937} as a kind of basic target detection algorithm to the corpus of categorized users described in Table \ref{table:users}. The objective was to obtain more relevant tweets for our under-populated classes (AGAINST and NEUTRAL in Catalan and NEUTRAL in Spanish). We manually revised the obtained topics and selected only those tweets which were clustered within the ``independence'' topic.


\begin{table}[!ht]
\centering
\begin{tabular}{lcc} \hline
CIC Corpus & Catalan & Spanish \\ \hline
Average tweet length (tokens)& 27.17 & 30.31 \\ \hline
\end{tabular}
\caption{Average tweet length in the Catalonia Independence Corpus.}\label{tab:length_ind_dataset}
\end{table}

Finally, we selected approximately 10,000 tweets (excluding those shorter than 4 words) per language keeping the proportion of users from the initial pool of crawled tweets. We split them keeping 60\% for training, and 20\% each for development and test. The average length of a tweet in the Catalan Independence Corpus (in Table \ref{tab:length_ind_dataset}) is slightly shorter than the average in the TW-1O dataset (see Table \ref{tab:length_tw_dataset}) given that our corpus does not include the previous and next tweets as context. However, our corpus is larger than previous works \cite{mohammad-etal-2016-semeval,taule18} and presents a more balanced distribution of classes, as shown by Table \ref{tab:distr_dataset}.

\begin{table}[!ht]
\centering
\begin{tabular}{lcc}\hline
Label & Catalan & Spanish \\ \hline
Against & 3988 & 4105 \\
Favor & 3902 & 4104 \\
Neutral & 2158 & 1868 \\ \hline
Total & 10048 & 10077 \\ \hline
\end{tabular}
\caption{Distribution of classes in the Catalonia Independence Corpus.}\label{tab:distr_dataset}
\end{table}

Finally, here we can see an example from the Catalonia Independence Corpus.

\begin{quote}
\textbf{Tweet:} \textit{Puigdemont visitar\`a el dia 13 de febrer la Universitat de Groningen dels Pa\"isos Baixos i presentar\'a  La crisi catalana, una oportunitat per Europa. \'Es un goig veure com ens reben els pa\"isos democr\`atics https://t.co/O38mDKwwn3}

\textbf{Stance:} FAVOR

\textbf{Language}: Catalan

\textbf{Translation}: \textit{Puigdemont will visit
on February 13th the University of Groningen, Netherlands, and present The Catalan Crisis, An Opportunity For Europe.
It's a pleasure to see how democratic countries are receiving us https://t.co/O38mDKwwn3}
\end{quote}

\subsection{Results}\label{sec:results}

This section reports on the results obtained by the systems presented in Section \ref{sec:experiment}.

\begin{table}[!ht]
\begin{tabular}{lccc}\hline
\textbf{System} & F1$_{against}$ & F1$_{favor}$ & F1$_{macro}$ \\ \hline
SVM+TF-IDF & 68.89 & 72.91 & 70.90 \\
SVM+FTEmb & 59.43 & 64.46 & 61.95 \\
fastText+FTEmb & 70.73 & 72.21 & \textbf{71.47} \\
Flair+FTEmb & 59.08 & 58.08 & 58.96 \\ \hline
\end{tabular}
\caption{Results on the Catalan testset of the Catalonia Independence Corpus (CIC-CA).}\label{tab:result_indep_ca}
\end{table}

\begin{table}[!ht]
\centering
\begin{tabular}{lccc}\hline
\textbf{System} & F1$_{against}$ & F1$_{favor}$ & F1$_{macro}$ \\ \hline
SVM+TF-IDF & 70.67 & 71.50 & 71.09 \\
SVM+FTEmb & 64.24 & 62.51 & 63.38 \\
fastText+FTEmb & 73.20 & 71.13 & \textbf{72.43} \\
Flair+FTEmb & 61.76 & 54.84 & 58.29 \\ \hline
\end{tabular}
\caption{Results on the Spanish testset of the Catalonia Independence Corpus (CIC-ES).}\label{tab:result_indep_es}
\end{table}

It is clear that the results for both Catalan (Table \ref{tab:result_indep_ca}) and Spanish (Table \ref{tab:result_indep_es}) are higher across languages and systems than those obtained on the TW-10 dataset. This means that the semi-automatic method for the annotation of tweets presented in this paper is quite effective and provides good quality annotated data. Furthermore, there is a consistency in the behaviour of the systems across both languages, which allows to compare their performance in multilingual settings. These results are also consistent with the TW-1O Spanish subset, where the fastText+FTEmb system also obtained the best scores. Finally, the deep learning approach from Flair seems to lag behind linear classifiers. While a bit surprising, this is also coherent with the results obtained in English with the SemEval 2016 dataset, as explained in the Related Work section. Our hypothesis is that the short length of the tweets make it more difficult to generate good contextual-based word representations. However, further experimentation is required to clarify this issue.

\section{Error Analysis}\label{sec:error}

In order to perform an analysis of the quality of the annotations obtained by our semi-automatic method (as described in Section \ref{sec:independence}), we took a sample of 100 tweets per language from the training sets. This sample was manually revised by three human annotators. We found out that the error rate in the Spanish sample was around 5\%, whereas for the Catalan sample was slightly higher, around 15\%. It should be noted that those error rates are approximate because the three human annotators found it very difficult to agree on their correct annotation. This was due to several reasons. First, the meaning of the tweets is usually underspecified. Second, many tweets use figurative language such as sarcasm and irony. Finally other tweets referred to the topic in a indirect manner. Below it can be found a couple of examples of contentious tweets in which it is not really clear which of the annotations are the correct one, namely, the one provided by our method (semi-automatic user-based) or the manual one. In Tweet 1, we can see a seemingly neutral message, but the author uses anti-independence slogan. In Tweet 2, although it seems to be neutral, the interpretation depends on the context of the message, where the annotator should know the details of the case.

\begin{quote}
\textbf{Tweet 1:} \emph{Arrimadas ir\'a a Waterloo este domingo para recordar a Puigdemont que  la rep\'ublica no existe  https://t.co/6luAEAj2UD}

\textbf{Our method:} NEUTRAL

\textbf{Manual annotation:} AGAINST/NEUTRAL

\textbf{Language:} Spanish

\textbf{Translation:} \emph{Arrimadas will go to Waterloo this Sunday to remind Puigdemont that the republic does not exist https://t.co/6luAEAj2UD}
\end{quote}

\begin{quote}
\textbf{Tweet 2:} \emph{@unprecisionman @jordisalvia Quan l'advocat preguntava sobre certes contradiccions d'un incident concret q havia explicat el Millo, el jutge ha dit q aix\`o no era rellevant per la causa}

\textbf{Our method:} AGAINST

\textbf{Manual annotation:} FAVOR/NEUTRAL

\textbf{Language:} Catalan

\textbf{Translation:} \emph{When the lawyer asked him about certain contradictions with respect to a specific incident which Millo had explained, the judge said that it was not relevant.}

\end{quote}

Manual annotation of stance in tweets is a difficult task for humans, partly because it depends greatly on the annotator's background knowledge and intuition. Furthermore, annotating tweets one by one, as opposed to user-based annotation, albeit automatic, suffers from a lack of context.





\section{Concluding Remarks}\label{sec:concluding-remarks}

In this paper we provide a new dataset for stance detection in Catalan and Spanish. The objective is two-fold: (i) to promote research on stance detection in other languages different to English and, (ii) to facilitate experimentation in multilingual and cross-lingual settings. We show that the methodology used to build the Catalonia Independence Corpus generates good quality annotated data without having to manually annotate tweet by tweet. Most importantly, it also helps to alleviate the imbalance in the classes distribution. Our experimental results confirm these considerations as the tested systems exhibit consistent behaviour across languages. We believe that our methodology can help to obtain larger annotated datasets from limited resources while making the annotation process cheaper and faster.

Additionally, we establish new state-of-the-art results on the TW-1O dataset for both Catalan and Spanish. Our hypothesis to explain the large difference with previous work is the more exhaustive pre-processing performed, apart from the use of the fastText word embeddings to obtain the tweets representation. The results also show the superior performance of the fastText linear classifier over SVM or RNN approaches on both datasets. These results are somewhat similar to those obtain for English with the SemEval 2016 data, where linear classifiers still are competitive or outperform newer deep learning approaches.

We publicly distribute the datasets and code to facilitate further multilingual and cross-lingual research on stance detection\footnote{https://github.com/ixa-ehu/catalonia-independence-corpus}.

\section{Acknowledgements}\label{sec:acknoledgements}

This work has been funded by the~Spanish Ministry of Science, Innovation and Universities under the project DeepReading (RTI2018-096846-B-C21) (MCIU/AEI/FEDER, UE) and by the BBVA Big Data 2018 ``BigKnowledge for Text Mining (BigKnowledge)'' project. The second author is funded by the Ramon y Cajal Fellowship RYC-2017-23647. We also acknowledge the~support of the Nvidia Corporation with the~donation of a Titan V GPU used for this research.

\section{Bibliographical References}\label{main:ref}

\bibliographystyle{lrec}
\bibliography{bibliography}


\end{document}